\documentclass[conference]{IEEEtran}
\IEEEoverridecommandlockouts

\usepackage{cite}
\usepackage{amsmath,amssymb,amsfonts}
\usepackage{algorithm}
\usepackage{algpseudocode}
\usepackage{graphicx}
\usepackage{textcomp}
\usepackage{xcolor}
\usepackage{booktabs}
\usepackage{float}

\def\BibTeX{{\rm B\kern-.05em{\sc i\kern-.025em b}\kern-.08em
    T\kern-.1667em\lower.7ex\hbox{E}\kern-.125emX}}
\begin{document}

\title{
Distribution-Free Pretraining of Classification Losses via Evolutionary Dynamics
}
\author{\IEEEauthorblockN{1\textsuperscript{st} Xiang Meng}
\IEEEauthorblockA{\textit{School of Computer Science and Engineering} \\
\textit{University of Aizu}\\
Aizuwakamatsu, Japan \\
d8242104@u-aizu.ac.jp}
\and
\IEEEauthorblockN{2\textsuperscript{nd} Yan Pei}
\IEEEauthorblockA{\textit{School of Computer Science and Engineering} \\
\textit{University of Aizu}\\
Aizuwakamatsu, Japan \\
peiyan@u-aizu.ac.jp}
}

\maketitle

\begin{abstract}
We propose Evolutionary Dynamic Loss (EDL), a framework that learns a transferable classification loss in probability space using unlimited synthetic prediction--label pairs, without accessing real samples during the main loss pretraining stage. 
EDL parameterizes the loss as a lightweight network and is trained with a semantics-free ranking-consistency objective that enforces larger penalties for more erroneous predictions. 
To robustly search loss shapes, we optimize EDL via an evolutionary strategy and introduce chaotic mutation to improve exploration under noisy fitness evaluation. 
Experiments on CIFAR-10 with ResNet backbones show that EDL can serve as a drop-in replacement for cross-entropy and achieves competitive or improved accuracy, while ablations confirm that chaotic mutation yields faster convergence and better synthetic-pretraining metrics than normal Gaussian mutation.
\end{abstract}


\section{Introduction}

Loss functions play a central role in supervised classification: they convert predicted probabilities into learning signals and thereby shape optimization dynamics, training stability, and generalization.
Despite their broad success, widely used objectives such as cross-entropy and its variants are fixed analytic forms.
When training conditions change---e.g., the hardness distribution, sampling bias, model capacity, label noise, or class imbalance---a fixed loss may no longer provide well-calibrated gradients or robust behavior.
This raises a fundamental question: can we learn a loss function that does not depend on any particular dataset, yet remains scalable and transferable across downstream settings?

Most existing approaches to learned losses or meta-learned objectives still rely on real task data and training trajectories as outer-loop supervision.
Consequently, these approaches face three limitations: supervision is bounded by dataset size and privacy constraints, dataset-specific biases can be absorbed into the learned objective and harm cross-task transfer, 
and learning a loss is closer to searching for a function shape than to fitting conventional parameters, making the optimization process noise-sensitive and prone to undesirable shapes that may destabilize downstream training.

In this work, we propose a loss-learning framework centered on two principles: distribution-free synthetic supervision in the probability space and evolutionary search for robust loss shapes.
Our key observation is that, for classification, a loss depends on the relationship between the predictive distribution and on the label rather than the raw input modality.
Therefore, we learn a parametric Evolutionary Dynamic Loss (EDL) directly in probability space by generating unlimited synthetic pairs $(p,y)$ without accessing any real samples.

\begin{figure*}[t]
  \centering
  \includegraphics[width=0.85\textwidth]{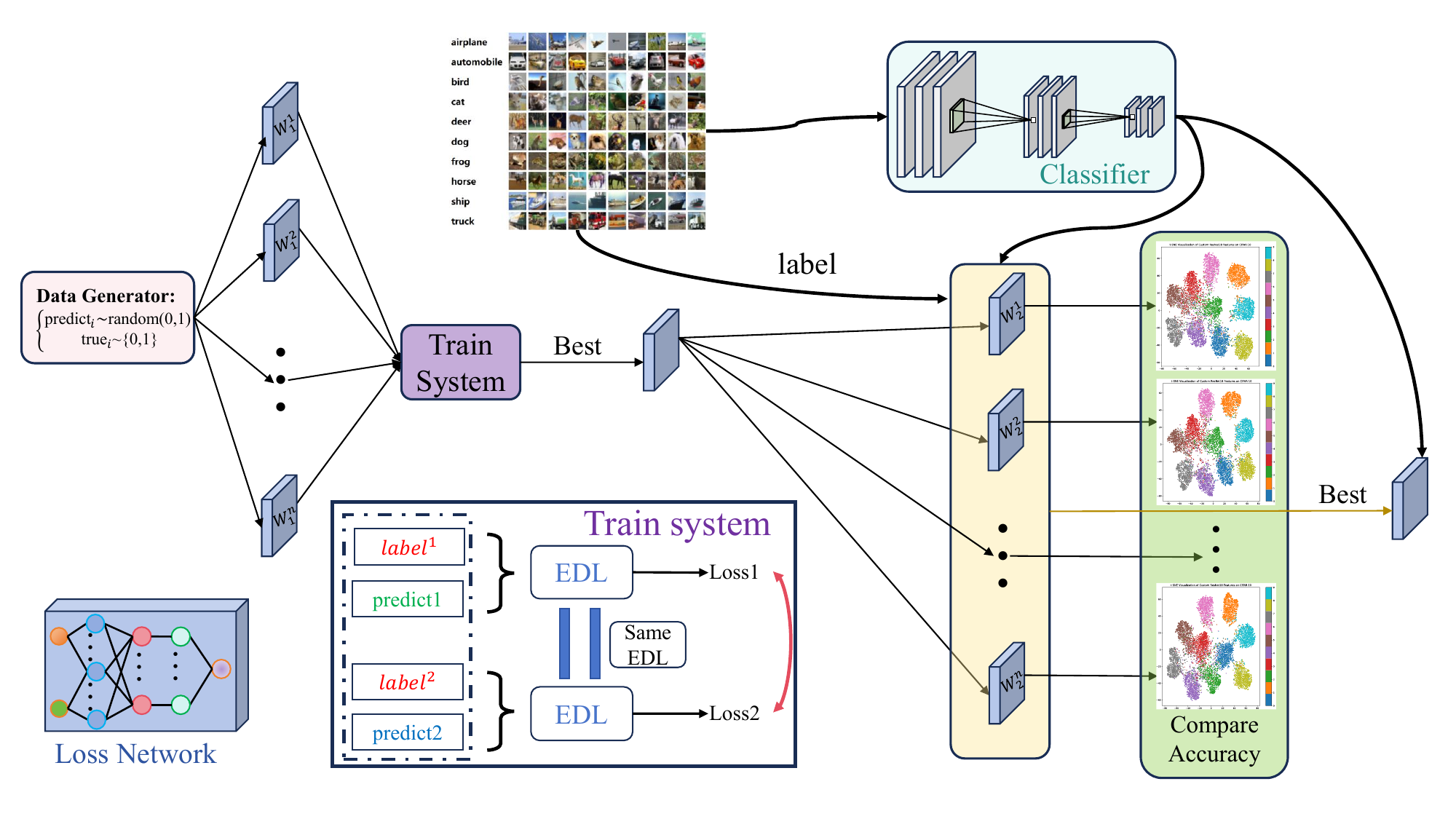}
  \caption{Overview of the proposed EDL pipeline.
  We learn a transferable loss prior from unlimited synthetic prediction--label pairs $(p,y)$ in probability space using a ranking-consistency signal and evolutionary search with chaotic mutation.
  The learned loss can then be used as a drop-in objective for standard downstream classifier training, optionally with a lightweight validation-based selection or calibration step on the target dataset.}
  \label{fig:edl_overview}
\end{figure*}

Figure~\ref{fig:edl_overview} summarizes the overall pipeline.
To obtain meaningful supervision without semantics, we impose a simple monotonicity principle: harder predictions should incur larger penalties.
Concretely, we sample synthetic pairs, define a deterministic hardness proxy (e.g., based on the true-class probability), and train EDL using a smooth pairwise ranking-consistency objective that encourages EDL to preserve the same ordering.
Moreover, we adopt a controllable mixture sampling scheme on the probability simplex to emphasize critical regimes that dominate training dynamics, including extreme-confidence cases and boundary-confusing predictions, which helps constrain the loss shape across the full confidence spectrum.

Optimizing such a shape-search objective can be brittle with purely gradient-based methods under noisy pairwise estimates.
We therefore employ an evolutionary strategy (ES) to explore EDL parameters globally.
In addition, we introduce Logistic-map-driven chaotic mutation to modulate mutation amplitudes in a bounded yet non-periodic manner, improving population diversity and reducing premature convergence in early search stages.
The outcome of this stage is a transferable loss prior learned without real data; for practical use, the learned loss can be plugged into standard classifier training as a drop-in objective, optionally followed by a lightweight validation-based selection step to better match a target dataset.

\noindent\textbf{Contributions.}
(i) We propose a distribution-free loss-learning formulation that learns a parametric loss directly from synthetic prediction--label pairs in probability space via a ranking-consistency objective.
(ii) We develop an evolutionary training strategy for loss-shape search, including Logistic-map-driven chaotic mutation, to improve global exploration and robustness under noisy pairwise supervision.
(iii) We demonstrate that the resulting loss prior can be plugged into standard training pipelines and yields stable and competitive generalization across downstream settings.

We will make our code and data publicly available upon acceptance.

\section{Related Work}

In supervised classification, performance can be improved by optimizing multiple components of the learning pipeline, including initialization~\cite{xiao2018dynamical}, learning-rate schedules~\cite{smith2019super}, activation functions~\cite{ramachandran2017searching}, data augmentation~\cite{wen2020time} and network architectures. 
Among these factors, the loss function is particularly influential: it defines the training signal and the error geometry, thereby shaping optimization dynamics, training stability, and generalization. 
Accordingly, recent work has increasingly focused on learning and adapting loss functions to better cope with task diversity and distributional shifts~\cite{gonzalez2020improved}.

\subsection{Static Loss Functions}
Static loss functions are specified prior to training and remain fixed throughout optimization. 
Classical objectives such as cross-entropy and hinge loss are widely adopted due to their simplicity and interpretability~\cite{NIPS1998_a14ac55a,zhang2004statistical}. 
A substantial line of research refines static losses by reshaping penalty profiles or introducing auxiliary terms to improve robustness and to enhance discriminative learning. 
However, because their functional forms do not change, static losses can be suboptimal when training conditions vary across stages or when the data distribution shifts, as they cannot dynamically reallocate learning pressure across different error regimes.

\subsection{Dynamic and Learnable Loss Functions}
To address the limited adaptability of static objectives, dynamic and learnable loss functions aim to adjust error evaluation during training based on the data distribution or the model state.
Representative approaches~\cite{wu2018learning} include learning adaptive coefficient structures that modulate the relationship between predictions and labels, meta-models~\cite{liu2020stochastic} that stochastically select and combine handcrafted losses, and meta-networks~\cite{baik2021meta} that generate transformation parameters to update a loss network.
Hai~\cite{hai2023l2t} further propose an LSTM-based meta-optimizer to update a loss network online, enabling stage-wise adaptation during training.
In addition, margin-based Softmax losses have been studied under unified formulations, where random search and policy-gradient methods~\cite{wang2018additive} are used to tune key hyperparameters for improved discriminative learning.
Overall, these methods highlight the promise of learning the optimization objective itself; however, they typically rely on real datasets and task-specific training trajectories to provide outer-loop supervision.

\paragraph{Evolutionary and gradient-free loss optimization.}
Complementary to trajectory-driven, gradient-based meta-learning, another line of work searches loss shapes using gradient-free optimization.
In particular, evolutionary computation can parameterize the loss and optimize it via mutation and selection, enabling global exploration without differentiability requirements.
Representative examples include Genetic Loss-function Optimization (GLO)~\cite{gonzalez2020improved} and its continuous parameterizations such as TaylorGLO~\cite{gonzalez2021taylorglo}, as well as evolutionary loss-function frameworks such as ELF~\cite{meng2025optimization}.
These perspectives motivate our use of evolutionary search to learn a transferable loss prior from probability-space supervision.

\section{Proposed Method}

We propose to learn a parametric loss function directly in probability space. 
Specifically, we model the loss as a evolutionary dynamic Loss that takes a classifier's predictive distribution and the ground-truth label as input and outputs a non-negative scalar loss. 
To make the learned loss scalable and dataset-agnostic, we construct synthetic supervision by sampling prediction--label pairs on the probability simplex, and optimize EDL parameters via an evolutionary strategy under a ranking-consistency objective. 
During evolution, we adopt chaotic mutation driven by a Logistic map to diversify the population and improve global exploration. 
After training, EDL can be plugged into standard training pipelines as the optimization objective for downstream classifiers.

\begin{figure}[t]
  \centering
  \includegraphics[width=0.9\linewidth, trim=1.4cm 0cm 1.8cm 0cm, clip]{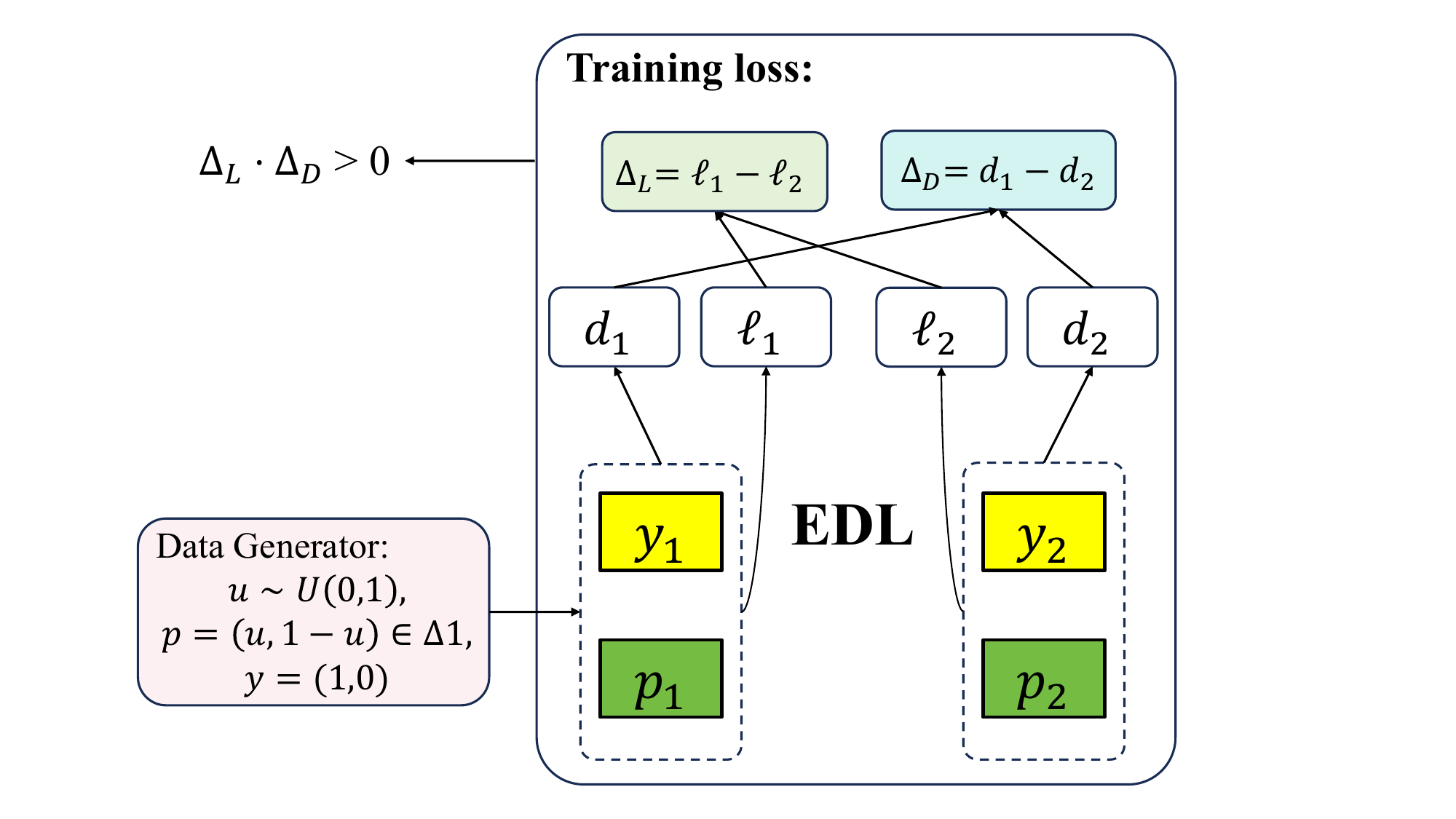}
  \caption{Synthetic supervision for pretraining EDL via ranking consistency in probability space.
  Two synthetic prediction--label pairs $(\mathbf p_1,\mathbf y_1)$ and $(\mathbf p_2,\mathbf y_2)$ are sampled on the probability simplex.
  EDL outputs scalar losses $\ell_1$ and $\ell_2$, and we form $\Delta_L=\ell_1-\ell_2$.
  A proxy hardness difference $\Delta_D$ is computed from the same pairs, and pretraining encourages the ordering agreement $\Delta_L\Delta_D>0$.}
  \label{fig:edl_stage1}
\end{figure}

Fig.~\ref{fig:edl_stage1} visualizes the core supervision signal used to train EDL without real data.
Note that the condition $\Delta L \cdot \Delta D > 0$ is not optimized as a discrete constraint; in practice we minimize a smooth pairwise ranking surrogate (e.g., softplus on $-\Delta L \cdot \Delta D$) to obtain stable gradients.
The proxy hardness is computed deterministically from $(\mathbf p,\mathbf y)$ (e.g., using the true-class probability or a distance between $\mathbf p$ and $\mathbf y$), and serves only to define a relative difficulty ordering between two synthetic samples.
This construction provides scalable supervision for shaping a monotone loss behavior across diverse confidence regimes, while remaining agnostic to any specific dataset.

\subsection{Evolutionary Dynamic Loss Network}
Let $z$ denote the logits produced by a classifier, and $p=\mathrm{softmax}(z)\in\Delta^{C-1}$ be the corresponding predictive distribution over $C$ classes. 
Given a label $y\in\{1,\dots,C\}$ with one-hot encoding $e_y\in\{0,1\}^{C}$, EDL defines a learned loss
\begin{equation}
\label{eq:EDL_def}
L_{\phi}(p,y)\;=\;\mathrm{EDL}_{\phi}\big([p,e_y]\big),
\end{equation}
where $[p,e_y]$ denotes concatenation and $\phi$ are EDL parameters. 
We enforce non-negativity by applying a Softplus output layer, ensuring $L_{\phi}(p,y)\ge 0$.

\begin{figure}[t]
  \centering
  \includegraphics[width=0.9\linewidth, trim=0.5cm 0.3cm 1cm 0cm, clip]{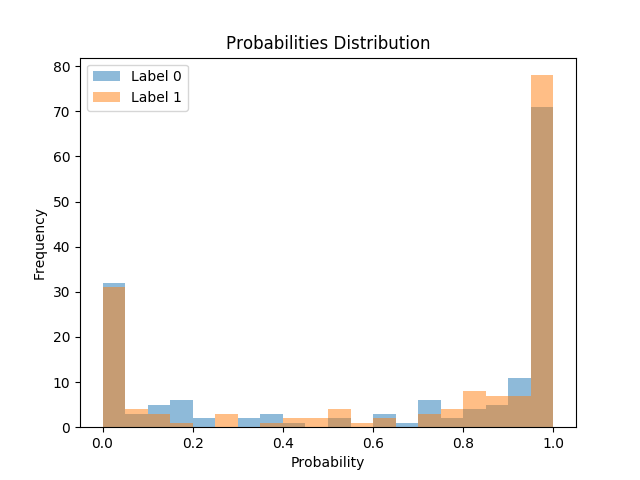}
  \caption{Histogram of synthetic probabilities used for EDL training. 
  We group samples by label and plot the distribution of predicted probabilities. 
  The mixture sampling intentionally covers both near-extreme regions (close to 0 or 1) and intermediate-confidence regions, providing diverse supervision for learning a ranking-consistent loss shape in probability space.}
  \label{fig:prob_dist_by_label}
\end{figure}

\subsection{Synthetic Data Construction in Probability Space}
To train EDL without using real samples, we generate synthetic training pairs $(p,y)$. 
We sample labels uniformly, $y\sim\mathrm{Unif}(\{1,\dots,C\})$, and draw $p$ from a controllable mixture on $\Delta^{C-1}$ to cover diverse confidence regimes. 
In particular, we strengthen two critical regions that frequently dominate training dynamics: 
(i) extreme-confidence predictions with probabilities close to $0$ or $1$, and 
(ii) boundary-confusing predictions where the true-class probability is close to that of a competing class. 
The resulting synthetic probabilities exhibit concentrated mass near the extremes while still maintaining broad coverage over intermediate-confidence ranges (see Fig.~\ref{fig:prob_dist_by_label}), providing scalable supervision for shaping the loss behavior across the full spectrum of prediction patterns.

\begin{algorithm}[H]
\caption{Training for EDL}
\label{alg:ce_EDL}
\begin{algorithmic}[1]
\Require Population size $K$, elite size $K_e$, generations $G$, batches per evaluation $B$, mutation scales $(\sigma_{\text{high}},\sigma_{\text{low}})$, threshold $\tau$, max attempts $A$
\Ensure Best EDL parameters $\phi^\star$
\State Initialize population $\mathcal{P}=\{\phi^{(k)}\}_{k=1}^{K}$; sample $x_0\sim\mathrm{Unif}(0,1)$; set $t\leftarrow 0$
\For{$g=1$ to $G$}
  \State Evaluate $\widehat{\mathcal{F}}(\phi^{(k)})$ for all $k$ using $B$ synthetic batches
  \State Select elites $\mathcal{E}\subset\mathcal{P}$ as the $K_e$ candidates with lowest $\widehat{\mathcal{F}}$; $\phi_{\text{best}}=\arg\min_{\phi\in\mathcal{E}} \widehat{\mathcal{F}}(\phi)$
  \State $\sigma \leftarrow \sigma_{\text{high}}$ if $\mathrm{Acc}(\phi_{\text{best}})<\tau$ else $\sigma_{\text{low}}$
  \State $\mathcal{P}' \leftarrow \mathcal{E}$
  \While{$|\mathcal{P}'|<K$}
    \State Sample parent $\phi\sim\mathrm{Unif}(\mathcal{E})$
    \For{$a=1$ to $A$}
      \State $x_{t+1} \leftarrow 4x_t(1-x_t)$; \ $t\leftarrow t+1$ 
      \State Sample $d\in\{-1,+1\}^{|\phi|}$ and $\epsilon\sim\mathcal{N}(0,I)$
      \State $\phi' \leftarrow \phi + \sigma\, x_{t+1}\,(d \odot \epsilon)$ \Comment{Chaotic mutation}
      \If{$\widehat{\mathcal{F}}(\phi')\le \widehat{\mathcal{F}}(\phi)$} \textbf{break} \EndIf
    \EndFor
    \State Add $\phi'$ to $\mathcal{P}'$
  \EndWhile
  \State $\mathcal{P} \leftarrow \mathcal{P}'$
\EndFor
\State \Return $\phi^\star=\arg\min_{\phi\in\mathcal{P}} \widehat{\mathcal{F}}(\phi)$
\end{algorithmic}
\end{algorithm}

\subsection{Ranking-Consistency Objective and Fitness}
Our supervision is based on a monotonicity principle: more incorrect predictions should incur larger loss values. 
For a synthetic sample $(p,y)$, we define an error-severity (hardness) score
\begin{equation}
\label{eq:hardness}
D(p,y)\;=\;1-p_y,
\end{equation}
where $p_y$ is the predicted probability assigned to the true class. 
We then sample pairs $(p_i,y_i)$ and $(p_j,y_j)$ and compare the ordering induced by $D$ and by EDL outputs. 
Let
\begin{equation}
\label{eq:pairwise_defs}
\begin{aligned}
\Delta D_{ij} &= D(p_i,y_i)-D(p_j,y_j), \qquad
s_{ij}=\mathrm{sign}(\Delta D_{ij}),\\
\Delta L_{ij} &= L_{\phi}(p_i,y_i)-L_{\phi}(p_j,y_j).
\end{aligned}
\end{equation}
The desired ranking consistency is $s\cdot \Delta L>0$. 
To obtain a stable and continuous training signal, we define the fitness as a smooth pairwise ranking loss:
\begin{equation}
\label{eq:fitness}
\mathcal{F}(\phi)
=\mathbb{E}_{(i,j)}\Big[\mathrm{softplus}\big(-s\cdot \Delta L\big)\Big],
\end{equation}
where the expectation is over randomly sampled pairs. 
Minimizing $\mathcal{F}(\phi)$ encourages EDL to assign larger loss to examples with larger error severity. 
For monitoring, we also report a trend-consistency accuracy:
\begin{equation}
\label{eq:acc}
\mathrm{Acc}(\phi)=\mathbb{P}_{(i,j)}\big[s\cdot \Delta L>0\big].
\end{equation}

\subsection{Evolutionary Optimization with Chaotic Mutation}
We optimize EDL parameters $\phi$ using a population-based evolutionary strategy over candidates $\mathcal{P}=\{\phi^{(k)}\}_{k=1}^{K}$. 
Since $\mathcal{F}(\phi)$ in Eq.~\ref{eq:fitness} is defined as an expectation over randomly sampled pairs, we evaluate each candidate by a Monte Carlo estimate computed on multiple synthetic batches:
\begin{equation}
\label{eq:es_fitness_est}
\begin{aligned}
\widehat{\mathcal{F}}\!\left(\phi^{(k)}\right)
&=\frac{1}{B}\sum_{b=1}^{B}\frac{1}{|\mathcal{S}_b|}
\sum_{(i,j)\in\mathcal{S}_b}
\mathrm{softplus}\!\Big(-s_{ij}\,\Delta L^{(k)}_{ij}\Big), \\
\Delta L^{(k)}_{ij}
&=L_{\phi^{(k)}}(p_i,y_i)-L_{\phi^{(k)}}(p_j,y_j).
\end{aligned}
\end{equation}
At each generation, we keep the top $K_e$ candidates with the lowest $\widehat{\mathcal{F}}$ as elites and generate the remaining candidates by chaotic mutation.

\paragraph{Logistic-map chaos.}
We use the Logistic map to generate a chaotic coefficient:
\begin{equation}
\label{eq:logistic_map}
x_{t+1} = r\,x_t(1-x_t),\qquad r=4.0,\; x_t\in(0,1).
\end{equation}

\paragraph{Chaotic mutation operator.}
Let $\theta_k$ denote the $k$-th learnable parameter in $\phi$. 
Given a mutation scale $\sigma$, we mutate each parameter by
\begin{equation}
\label{eq:chaotic_mutation}
\theta_k'=\theta_k+\sigma\, d_k\, x_{t+1}\,\epsilon_k,\qquad 
d_k\in\{-1,+1\},\;\epsilon_k\sim\mathcal{N}(0,1),
\end{equation}
where $d_k$ controls mutation direction and the chaotic coefficient $x_{t+1}$ adaptively modulates the step size in a bounded yet non-linear manner. 
Compared with a fixed-scale Gaussian mutation, chaotic mutation yields richer perturbation patterns and improves early-stage exploration of diverse loss shapes.

We use an adaptive mutation scale (switching between $\sigma_{\text{high}}$ and $\sigma_{\text{low}}$ based on the best $\mathrm{Acc}$) to balance exploration and refinement. 
The overall procedure is summarized in Algorithm~\ref{alg:ce_EDL}.

\subsection{Algorithm Explanation}
Algorithm~\ref{alg:ce_EDL} repeats evaluation--selection--chaotic mutation: it evaluates candidates by $\widehat{\mathcal{F}}$, keeps the best $K_e$ elites, and refills the population using Logistic-map-driven chaotic mutation with an adaptive $\sigma$. 
The final output is the EDL with the lowest estimated fitness.

\subsection{Theoretical Analysis}

\paragraph{Contrastive supervision in probability space}
EDL is trained in probability space using synthetic pairs $(p,y)$, which is sufficient because a classification loss depends only on the relation between predictive probabilities and labels.
We reuse the hardness score $D(p,y)$ in Eq.\eqref{eq:hardness}, the pairwise ordering variables $(\Delta D_{ij}, s_{ij}, \Delta L_{ij})$ in Eq.\eqref{eq:pairwise_defs}, and the ranking-consistency objective $\mathcal{F}(\phi)$ in Eq.\eqref{eq:fitness}.
Minimizing Eq.\eqref{eq:fitness} enforces the monotonicity principle that harder predictions (larger $D$) should incur larger loss values (larger $L_\phi$).
Moreover, if there exists a strictly increasing $g(\cdot)$ such that $L^\star(p,y)=g(D(p,y))$, then $\mathrm{sign}(\Delta L^\star_{ij})=\mathrm{sign}(\Delta D_{ij})$ for any $(i,j)$; hence the objective in Eq.\eqref{eq:fitness} directly promotes a monotone loss shape with a positive ordering margin.

\paragraph{Extreme coverage via mixture sampling}
Uniform sampling on the simplex may under-represent extreme-confidence regimes ($p_y\!\approx\!0$ or $1$), which are often the most sensitive regions for gradient magnitudes and training dynamics.
Our synthetic construction therefore draws $(p,y)$ from a controllable mixture distribution (see Fig.~\ref{fig:prob_dist_by_label}), which increases the fraction of pairwise constraints involving near-extreme samples.
This reduces extrapolation uncertainty at the boundaries and improves ordering consistency across the full confidence spectrum induced by $D(p,y)$.

\paragraph{Stability under stochastic fitness evaluation}
Because $\mathcal{F}(\phi)$ is an expectation over randomly sampled pairs, the ES fitness estimate $\widehat{\mathcal{F}}(\phi)$ in Eq.\eqref{eq:es_fitness_est} is noisy.
Averaging over $B$ independent batches reduces estimator variance approximately inversely with $B$:
\begin{equation}
\label{eq:theory_batchavg_var}
\widehat{\mathcal{F}}(\phi)=\frac{1}{B}\sum_{b=1}^{B}\widehat{\mathcal{F}}_{b}(\phi),
\qquad
\mathrm{Var}\!\left[\widehat{\mathcal{F}}(\phi)\right]\approx \frac{1}{B}\mathrm{Var}\!\left[\widehat{\mathcal{F}}_{1}(\phi)\right].
\end{equation}
Elitism further protects the current best candidate, and the optional non-degradation acceptance rule
\begin{equation}
\label{eq:theory_accept}
\widehat{\mathcal{F}}(\phi')\le \widehat{\mathcal{F}}(\phi)
\end{equation}
reduces random drift caused by unlucky perturbations under noisy evaluation.

\paragraph{Why chaotic mutation helps (bounded, intermittent steps)}
Normal mutation uses fixed-scale Gaussian perturbations, while our chaotic mutation in Eq.\eqref{eq:chaotic_mutation} introduces a bounded, time-varying scale factor generated by the Logistic map in Eq.\eqref{eq:logistic_map}.
Equivalently, each mutation step can be viewed as
\begin{equation}
\label{eq:theory_step}
\Delta\theta=\sigma\,x\,\epsilon,\qquad x\in(0,1),\;\epsilon\sim\mathcal{N}(0,1),
\end{equation}
where $x$ follows the chaotic dynamics induced by Eq.\eqref{eq:logistic_map}.
In the canonical chaotic regime, $x$ is ergodic on $(0,1)$ with an invariant distribution that places substantial mass near both $0$ and $1$, yielding an intermittent step pattern: most mutations are strongly down-scaled (small steps) while a non-negligible fraction are near unscaled (large steps).
This creates a natural ``many-small, few-large'' exploration behavior: small steps stabilize selection by reducing sensitivity to noisy fitness estimates, whereas rare large steps help escape poor local loss shapes.
Combined with elitist selection and the acceptance rule in Eq.\eqref{eq:theory_accept}, chaotic mutation provides a simple mechanism to balance exploitation and exploration during loss-shape search.

\section{Experimental Results}
\label{sec:exp}

\begin{table}[t]
\centering
\caption{Implementation details (EDL). Stage~1 synthetic pretraining is the focus.}
\label{tab:impl_summary_edl}
\scriptsize
\setlength{\tabcolsep}{5pt}
\renewcommand{\arraystretch}{1.05}
\begin{tabular}{l l}
\toprule
\multicolumn{2}{c}{\textbf{Downstream (CIFAR-10)}}\\
\midrule
Backbone / epochs         & ResNet / 200 \\
Optimiser / batch         & SGD (momentum 0.9, wd $5\times 10^{-4}$) / 128 \\
LR schedule               & init $10^{-2}$, step $\times0.1$ at $\{120,160\}$ \\
Loss                      & CE or EDL \\
\midrule
\multicolumn{2}{c}{\textbf{Stage~1: Synthetic pretraining (probability space)}}\\
\midrule
Classes / input           & $C=10$; input: predicted prob. vector + one-hot label \\
Loss network              & MLP (2C-10-20-20-1) + Softplus \\
Ranking supervision       & Pairwise ranking on synthetic samples (Softplus ranking loss) \\
Synthetic budget          & $A=8192$ samples, $B=4096$ pairs per generation \\
\midrule
\multicolumn{2}{c}{\textbf{Optimisation (ES)}}\\
\midrule
Population / elites       & $K=6$, $K_e=2$ \\
Generations               & 80 \\
Mutation (Normal)         & Gaussian perturbation \\
Mutation (Chaotic)        & Logistic-modulated perturbation (chaos factor $x_0\sim U(0,1)$) \\
Noise schedule            & threshold $\tau=0.95$; $\sigma_{\text{high}}=0.20$, $\sigma_{\text{low}}=0.01$ \\
\bottomrule
\end{tabular}
\end{table}

\subsection{Experimental Setup}

\begin{table}[ht]
\centering
\caption{Results on datasets CIFAR-10~\cite{krizhevsky2009learning} for the classification task. All experiments are implemented with the same settings. The best results are highlighted in bold.}
\resizebox{0.95\linewidth}{!}{%
\begin{tabular}{|l|ccc|}
\hline
Method & \multicolumn{3}{c|}{CIFAR-10~\cite{krizhevsky2009learning}} \\ \cline{2-4}
       & ResNet8   & ResNet20  & ResNet32  \\ \hline
CE     & 87.6      & 91.3      & 92.5      \\
Smooth~\cite{nguyen2013algorithms} & 87.9 & 91.5 & 92.6 \\
L-M Softmax~\cite{liu2016large} & 88.7 & 92.0 & 93.0 \\
L2T-DLF~\cite{wu2018learning} & 89.2 & 92.4 & 93.1 \\
GLO~\cite{gonzalez2020improved} & 87.7 & 91.3 & - \\
TaylorGLO~\cite{gonzalez2021taylorglo} & - & 91.6 & - \\
ARLF~\cite{barron2019general} & 89.5 & 91.5 & 92.2 \\
SLF~\cite{liu2020stochastic} & 89.8 & 93.0 & 93.9 \\
ALA~\cite{huang2019addressing} & - & - & 93.2 \\
L2T-DLN~\cite{hai2023l2t} &  90.7 & 93.4 & 93.8 \\
ELF~\cite{meng2025optimization} & 90.4  & 92.9 & 93.0\\
EDL-GD & 90.6 $\pm$ 0.18 & 91.9 $\pm$ 0.08 & 92.3 $\pm$ 0.24 \\
EDL-ES-Chaotic & \textbf{91.1 $\pm$ 0.14} & \textbf{93.4 $\pm$ 0.17} & \textbf{93.9 $\pm$ 0.12}\\ \hline 
\end{tabular}%
}
\label{tab:cifar10_comparison}
\end{table}

\paragraph{Datasets}
We evaluate EDL on CIFAR-10~\cite{krizhevsky2009learning}, which contains 50000 training images and 10000 test images of size $32\times32$.

\paragraph{Implementation details}
Our method contains two learning components: (1) synthetic pretraining of the parametric loss in probability space, and (2) downstream classifier training using the learned loss.
Table~\ref{tab:impl_summary_edl} summarizes the shared architectures and hyper-parameters.
For the classifier, we use the standard ResNet backbone.
For EDL, we use a lightweight MLP that maps the concatenated input $[p; e_y]$ to a non-negative scalar via a Softplus head, following the same loss-network architecture as~\cite{meng2025optimization}.
In the evolutionary setting, we maintain a population of candidate losses, keep elites, and generate offspring by mutation; chaotic mutation uses a Logistic map to modulate the mutation amplitude.
Experiments run on a single NVIDIA GPU (RTX3060). 

\paragraph{Evaluation metric}
Downstream performance is measured by Top-1 accuracy (\%) on the CIFAR-10 test set.
For the synthetic pretraining process, we report the best (lowest) fitness $\widehat{\mathcal{F}}$ and the trend-consistency accuracy $\mathrm{Acc}$.

\subsection{Results}
\label{sec:results}

\paragraph{Synthetic pretraining behaviour.}
We first examine whether synthetic supervision can shape a meaningful loss prior.
Across runs, EDL trained with evolutionary strategy exhibits steadily decreasing fitness and increasing trend-consistency accuracy, indicating that the learned loss preserves the desired monotonic ordering (harder predictions incur larger loss).
Compared with Gaussian mutation, chaotic mutation improves population diversity and typically finds better loss candidates with more stable convergence.

\paragraph{Downstream CIFAR-10 classification.}
We then plug the pretrained EDL into a standard ResNet training pipeline on CIFAR-10.
Table~\ref{tab:cifar10_comparison} summarizes the test accuracy.
EDL-ES-Chaos yields the best accuracy among learned-loss variants and improves over EDL-ES-Normal, demonstrating the benefit of chaotic exploration during loss pretraining.
In contrast, EDL-GD, while achieving high synthetic ranking consistency, transfers less effectively to CIFAR-10, suggesting that directly optimizing EDL by gradient descent on synthetic pairs may lead to less robust loss shapes for real-data optimization.

\begin{table}[ht]
\centering
\caption{Normal vs.\ chaotic mutation in ES-based EDL synthetic pretraining (fitness $\downarrow$, best\_acc $\uparrow$).}
\label{tab:chaos_vs_normal}
\scriptsize
\setlength{\tabcolsep}{2.5pt}
\renewcommand{\arraystretch}{1.05}
\resizebox{\columnwidth}{!}{%
\begin{tabular}{lcccccc}
\toprule
Mutation &
\shortstack{Final\\best $\downarrow$} &
\shortstack{Mean\\fit $\downarrow$} &
\shortstack{Max\\acc $\uparrow$} &
\shortstack{Mean\\acc $\uparrow$} &
\shortstack{Std\\acc $\downarrow$} &
\shortstack{Epoch@\\best $\downarrow$} \\
\midrule
Normal  & 0.02810 & 0.25742 & 100.00 & 99.18 & 1.70 & 80 \\
Chaotic & \textbf{0.01994} & \textbf{0.15528} & 100.00 & \textbf{99.60} & \textbf{1.44} & \textbf{78} \\
\midrule
Gain & $\mathbf{-29.05}\%$ & $\mathbf{-39.68}\%$ & $+0.00\%$ & $+0.43$ & $\mathbf{-15.32}\%$ & $-2$ \\
\bottomrule
\end{tabular}%
}
\end{table}

\subsection{Chaotic vs.\ Normal Mutation ablation experiments}
\label{sec:ablation_stage1_mutation}

To isolate the effect of the mutation operator in synthetic EDL pretraining, we compare chaotic mutation (Eq.~\ref{eq:logistic_map}--Eq.~\ref{eq:chaotic_mutation}) with normal Gaussian mutation under identical evolutionary settings.
We keep the ES population size, elite selection, evaluation budget (generations and Monte Carlo batches), acceptance rule, and the adaptive mutation-scale schedule fixed; the only difference is whether the mutation amplitude is modulated by a chaotic coefficient (Chaotic) or uses the standard Gaussian perturbation with the same scale (Normal).

\begin{figure}[H]
  \centering
  \includegraphics[width=\linewidth, trim=1.2cm 0cm 0cm 0cm, clip]{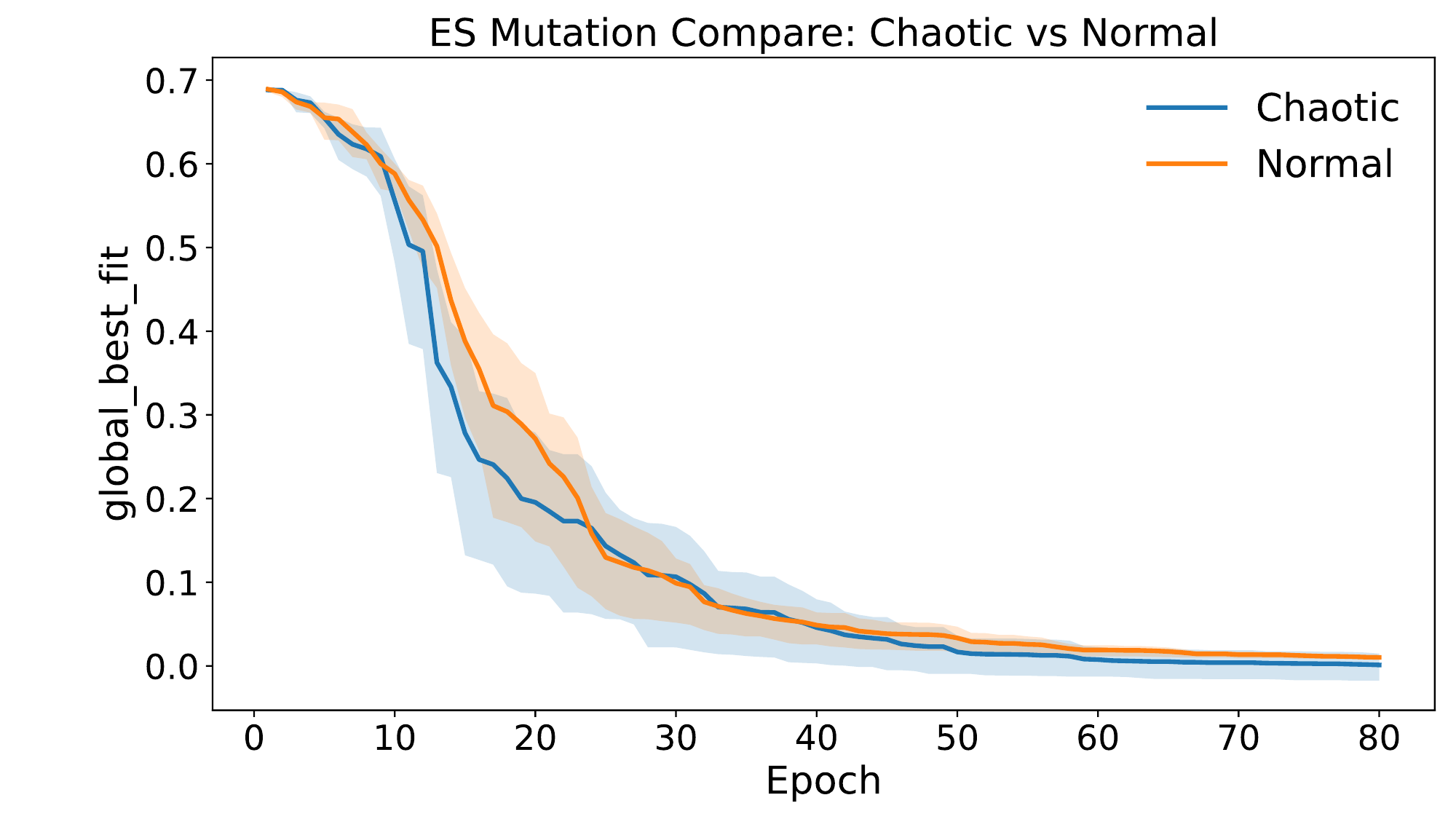}
  \caption{Ablation on the mutation operator in ES-based EDL pretraining.
  We plot global\_best\_fit (best-so-far synthetic ranking fitness; lower is better) over epochs, shown as mean $\pm$ std across seeds.
  Compared with normal Gaussian mutation, chaotic mutation reduces the fitness faster in the early-to-mid stage and achieves a slightly better final global-best fitness under the same ES budget, suggesting improved exploration and more robust loss-shape search.}
  \label{fig:stage1_chaos_vs_normal_curve}
\end{figure}

Fig.~\ref{fig:stage1_chaos_vs_normal_curve} shows the trajectory of \textit{global\_best\_fit} during synthetic pretraining, where \textit{global\_best\_fit} denotes the best-so-far ranking-consistency fitness $\widehat{\mathcal{F}}$ found up to each epoch (lower is better); curves are reported as mean $\pm$ std across seeds.
Both variants steadily reduce $\widehat{\mathcal{F}}$, confirming that probability-space supervision can shape a meaningful loss prior.
Notably, chaotic mutation attains lower fitness earlier in the search (early-to-mid epochs), while both methods become close as they converge.
This early advantage translates into a lower average fitness over epochs and a better best-so-far solution overall.

Table~\ref{tab:chaos_vs_normal} quantitatively corroborates this observation.
Compared with normal mutation, chaotic mutation achieves a substantially lower final global-best fitness (0.01994 vs.\ 0.02810) and a much lower mean fitness across epochs, while maintaining saturated ranking accuracy (max best\_acc $=100\%$) with improved stability (lower std. best\_acc) and slightly faster attainment of the best solution.

\section{Conclusion}
We presented Evolutionary Dynamic Loss (EDL), a distribution-free loss pretraining framework that learns a transferable loss prior from unlimited synthetic prediction--label pairs $(p,y)$ in probability space, without accessing real samples during the main pretraining stage. The loss is identified via a semantics-free ranking-consistency objective that enforces a monotone penalty ordering with respect to prediction hardness, which is directly reflected by steadily improved synthetic ranking fitness and consistency. To make the shape search robust under stochastic fitness evaluation, we optimize EDL with an evolutionary strategy and a Logistic-map-driven chaotic mutation that accelerates early-to-mid progress and improves aggregated pretraining metrics under the same ES budget; the resulting loss can then be plugged into standard CIFAR-10 training as a drop-in replacement and achieves competitive or improved Top-1 accuracy compared with strong learned-loss baselines.

\bibliographystyle{IEEEtran}
\bibliography{ref}

@inproceedings{gonzalez2020improved,
  title={Improved training speed, accuracy, and data utilization through loss function optimization},
  author={Gonzalez, Santiago and Miikkulainen, Risto},
  booktitle={2020 IEEE congress on evolutionary computation (CEC)},
  pages={1--8},
  year={2020},
  organization={IEEE}
}

@inproceedings{NIPS1998_a14ac55a,
 author = {Gentile, Claudio and Warmuth, Manfred K. K},
 booktitle = {Advances in Neural Information Processing Systems},
 editor = {M. Kearns and S. Solla and D. Cohn},
 pages = {},
 publisher = {MIT Press},
 title = {Linear Hinge Loss and Average Margin},
 volume = {11},
 year = {1998}
}

@article{zhang2004statistical,
  title={Statistical behavior and consistency of classification methods based on convex risk minimization},
  author={Zhang, Tong},
  journal={The Annals of Statistics},
  volume={32},
  number={1},
  pages={56--85},
  year={2004},
  publisher={Institute of Mathematical Statistics}
}

@inproceedings{barron2019general,
  title={A general and adaptive robust loss function},
  author={Barron, Jonathan T},
  booktitle={Proceedings of the IEEE/CVF conference on computer vision and pattern recognition},
  pages={4331--4339},
  year={2019}
}

@inproceedings{xiao2018dynamical,
  title={Dynamical isometry and a mean field theory of cnns: How to train 10,000-layer vanilla convolutional neural networks},
  author={Xiao, Lechao and Bahri, Yasaman and Sohl-Dickstein, Jascha and Schoenholz, Samuel and Pennington, Jeffrey},
  booktitle={International Conference on Machine Learning},
  pages={5393--5402},
  year={2018},
  organization={PMLR}
}

@inproceedings{smith2019super,
  title={Super-convergence: Very fast training of neural networks using large learning rates},
  author={Smith, Leslie N and Topin, Nicholay},
  booktitle={Artificial intelligence and machine learning for multi-domain operations applications},
  volume={11006},
  pages={369--386},
  year={2019},
  organization={SPIE}
}

@article{wen2020time,
  title={Time series data augmentation for deep learning: A survey},
  author={Wen, Qingsong and Sun, Liang and Yang, Fan and Song, Xiaomin and Gao, Jingkun and Wang, Xue and Xu, Huan},
  journal={arXiv preprint arXiv:2002.12478},
  year={2020}
}

@article{wu2018learning,
  title={Learning to teach with dynamic loss functions},
  author={Wu, Lijun and Tian, Fei and Xia, Yingce and Fan, Yang and Qin, Tao and Jian-Huang, Lai and Liu, Tie-Yan},
  journal={Advances in neural information processing systems},
  volume={31},
  year={2018}
}

@inproceedings{liu2020stochastic,
  title={Stochastic loss function},
  author={Liu, Qingliang and Lai, Jinmei},
  booktitle={Proceedings of the AAAI Conference on Artificial Intelligence},
  volume={34},
  number={04},
  pages={4884--4891},
  year={2020}
}

@inproceedings{huang2019addressing,
  title={Addressing the loss-metric mismatch with adaptive loss alignment},
  author={Huang, Chen and Zhai, Shuangfei and Talbott, Walter and Martin, Miguel Bautista and Sun, Shih-Yu and Guestrin, Carlos and Susskind, Josh},
  booktitle={International conference on machine learning},
  pages={2891--2900},
  year={2019},
  organization={PMLR}
}

@inproceedings{baik2021meta,
  title={Meta-Learning with Task-Adaptive Loss Function for Few-Shot Learning},
  author={Baik, Sungyong and Choi, Janghoon and Kim, Heewon and Cho, Dohee and Min, Jaesik and Lee, Kyoung Mu},
  booktitle={Proceedings of the IEEE/CVF International Conference on Computer Vision},
  pages={9465--9474},
  year={2021}
}

@inproceedings{hai2023l2t,
  title={L2T-DLN: Learning to Teach with Dynamic Loss Network},
  author={Hai, Zhaoyang and Pan, Liyuan and Liu, Xiabi and Liu, Zhengzheng and Yunita, Mirna},
  booktitle={Thirty-seventh Conference on Neural Information Processing Systems},
  year={2023}
}

@article{ramachandran2017searching,
  title={Searching for activation functions},
  author={Ramachandran, Prajit and Zoph, Barret and Le, Quoc V},
  journal={arXiv preprint arXiv:1710.05941},
  year={2017}
}

@article{wang2018additive,
  title={Additive margin softmax for face verification},
  author={Wang, Feng and Cheng, Jian and Liu, Weiyang and Liu, Haijun},
  journal={IEEE Signal Processing Letters},
  volume={25},
  number={7},
  pages={926--930},
  year={2018},
  publisher={IEEE}
}

@inproceedings{meng2025optimization,
  title={Optimization Design of Adaptive Loss Function Using Evolutionary Neural Networks},
  author={Meng, Xiang and Hai, Zhaoyang and Liu, Xiabi and Pei, Yan},
  booktitle={International Conference on Neural Information Processing},
  pages={321--335},
  year={2025},
  organization={Springer}
}

@article{krizhevsky2009learning,
  title={Learning multiple layers of features from tiny images},
  author={Krizhevsky, Alex and Hinton, Geoffrey and others},
  year={2009},
  publisher={Toronto, ON, Canada}
}

@inproceedings{nguyen2013algorithms,
  title={Algorithms for direct 0--1 loss optimization in binary classification},
  author={Nguyen, Tan and Sanner, Scott},
  booktitle={International conference on machine learning},
  pages={1085--1093},
  year={2013},
  organization={PMLR}
}

@article{liu2016large,
  title={Large-margin softmax loss for convolutional neural networks},
  author={Liu, Weiyang and Wen, Yandong and Yu, Zhiding and Yang, Meng},
  journal={arXiv preprint arXiv:1612.02295},
  year={2016}
}

@inproceedings{gonzalez2021taylorglo,
  title={Improved training speed, accuracy, and data utilization through loss function optimization},
  author={Gonzalez, Santiago and Miikkulainen, Risto},
  booktitle={2020 IEEE congress on evolutionary computation (CEC)},
  pages={1--8},
  year={2020},
  organization={IEEE}
}

\end{document}